%% file: iclr2026_conference.tex
\documentclass[11pt]{article} % For LaTeX2e
\usepackage[preprint]{acl}

\input{math_commands.tex}

\usepackage{hyperref}
\usepackage{url}
\usepackage{times}
\usepackage{latexsym}
\usepackage{amsmath}
\usepackage{amssymb}
\usepackage{algorithm}
\usepackage{algorithmic}
\usepackage{adjustbox} % Add to preamble

\usepackage{booktabs}  % For professional-quality tables with \toprule, \midrule, etc.
\usepackage{multirow}  % For the \multirow command (used for the 'Model' row spanning two header rows)
\usepackage{nccmath}  % for \smaller text in math
\usepackage{mathtools} % for better spacing control
\usepackage{tabularx}
\usepackage[most]{tcolorbox}
\usepackage[dvipsnames]{xcolor}
\usepackage{enumitem}
\usepackage[T1]{fontenc}

\usepackage[utf8]{inputenc}
\usepackage{microtype}
\usepackage{inconsolata}
\usepackage{graphicx}
\usepackage{ragged2e}
\newcolumntype{L}{>{\RaggedRight\arraybackslash}X}
\newcommand{\qa}[3][]{% #1 optional small note, #2 question, #3 answer
  \noindent\begin{tabularx}{\linewidth}{@{}L>{\raggedleft\arraybackslash}m{0pt}@{}}
    \textbf{Q: } #2 & \\[-.25ex]
    & \makebox[0pt][r]{\texttt{#3}}\\[-.25ex]
    \multicolumn{2}{@{}L}{\if\relax\detokenize{#1}\relax\else{\small #1}\fi}\\[.6ex]
  \end{tabularx}
}

\title{Cross-Examination Framework: A Task-Agnostic Diagnostic for Information Fidelity in Text-to-Text Generation}
\author{Tathagata Raha, Cl{\'e}ment Christophe, Nada Saadi, Hamza A Javed, \\
\textbf{Marco AF Pimentel, Ronnie Rajan, Praveenkumar Kanithi} \\
M42 Health, Abu Dhabi, UAE\\
\texttt{traha@m42.ae}}

\begin{document}

\maketitle

\begin{abstract}

Traditional metrics like BLEU and BERTScore fail to capture semantic fidelity in generative text-to-text tasks. We adapt the Cross-Examination Framework (CEF) for a reference-free, multi-dimensional evaluation by treating the source and candidate as independent knowledge bases. CEF generates verifiable questions from each text and performs a cross-examination to derive three interpretable scores: Coverage, Conformity, and Consistency. Validated across translation, summarization and clinical note-generation, our framework identifies critical errors, such as content omissions and factual contradictions, missed by standard metrics. A key contribution is a systematic robustness analysis to select a stable judge model. Crucially, the strong correlation between our reference-free and with-reference modes validates CEF's reliability without gold references. Furthermore, human expert validation demonstrates that CEF mismatching questions align with meaning-altering semantic errors higher than with non-semantic errors, particularly excelling at identifying entity-based and relational distortions.

\end{abstract}

\section{Introduction}
Evaluating generative text-to-text tasks requires assessing information fidelity: the degree to which a generated document preserves the source's meaning without omissions, distortions, or hallucinations. Whether in machine translation (MT), abstractive summarization, or structured note generation, the core challenge remains ensuring that the semantic content of the output is factually grounded in the source. Current metrics fall short: overlap-based methods like BLEU~\citep{papineni-etal-2002-bleu} penalize valid paraphrasing, while embedding-based metrics like BERTScore~\citep{Zhang2019BERTScoreET} are insensitive to factual errors. Furthermore, their heavy reliance on costly human references limits their applicability in domain-specific settings, misaligning evaluation with the core goal of semantic faithfulness.

To address these gaps, we adapt the Cross-Examination Framework (CEF)~\citep{kanithi2024medic}, as a task-agnostic, reference-free approach based on a simple intuition: if two texts contain the same information, they should answer the same questions. Originally introduced for clinical summarization, we generalize CEF as a universal diagnostic tool for text-to-text generation. CEF treats the source and generated output as independent knowledge bases, generates verifiable closed-ended questions from both, and performs a cross-examination. This process yields three interpretable scores, Coverage, Conformity and Consistency, that measure content retention, contradiction, and hallucination. We validate the generalizability of the framework across three diverse tasks: machine translation (NTREX dataset), abstractive summarization (CNN/DailyMail), and clinical note generation (ACI-Bench) in both reference-free and with-reference settings.

Our experiments show that CEF correlates with standard metrics while capturing critical failure modes, including factual contradictions and omissions, that are often overlooked, particularly in morphologically rich languages where conventional metrics degrade. The main contributions of this work are:
\begin{itemize}
\item \textbf{Task-Agnostic Framework:} We establish CEF as a reference-free, multi-dimensional assessment of information fidelity applicable across translation, summarization, and clinical tasks. 
\item \textbf{Multi-Task Validation:} Empirical validation on the NTREX, CNN/DM, and ACI-BENCH datasets, demonstrating that CEF uncovers semantic errors missed by traditional overlap and embedding metrics
\item \textbf{Robustness \& Reliability Analysis:} A systematic study of judge model stability and question count, establishing $N=10$ as a near-optimal configuration for balancing reliability with efficiency
\item \textbf{Reproducibility:} Public release of prompts, code, and detailed ablation studies to support reproducibility.
\end{itemize}

\section {Related Work}

Traditional metrics like BLEU and ROUGE rely on n-gram overlap and suffer from well-documented deficiencies. They exhibit surface-form bias by penalizing valid paraphrasing \citep{reiter-2018-structured} and, crucially, fail to correlate with factual consistency as demonstrated in human-centric evaluations of summarization and translation \citep{maynez-etal-2020-faithfulness, falke-etal-2019-ranking}. While embedding-based metrics like BERTScore \citep{zhang-etal-2020-bertscore} and MoverScore \citep{zhao-etal-2019-moverscore} mitigate some lexical issues, they remain largely insensitive to critical factual contradictions and entity-level hallucinations \citep{pagnoni-etal-2021-understanding, kryscinski-etal-2020-evaluating}. All these approaches are fundamentally limited by their dependency on gold-standard human references, as originally required by their scoring formulations \citep{papineni-etal-2002-bleu, lin-2004-rouge}.

Question-answering (QA) evaluation was introduced to address these semantic gaps. Early methods like QAEval \citep{deutsch-etal-2021-towards} were more flexible than n-gram overlap but remained tied to references. Subsequent work such as FEQA \citep{durmus-etal-2020-feqa} and QAGS \citep{wang-etal-2020-asking} pioneered reference-free evaluation by generating questions from the system output to check against the source. This was generalized by QuestEval \citep{scialom-etal-2021-questeval}, which uses a bidirectional approach to measure both coverage and consistency. SummEQuAL \cite{liu-etal-2024-summequal} further refined this QA paradigm specifically for summarization by using LLMs to gauge both recall and precision through content-based comparison. Despite these advances, existing QA-based methods often suffer from error propagation in their multi-step pipelines and have struggled to reliably outperform simpler metrics, highlighting challenges in robustness \citep{scialom-etal-2021-questeval}. 

To address the limitations of static overlap and error propagation, the field has recently shifted toward high-capacity neural evaluators and LLM-as-judge frameworks. Metrics like xCOMET \citep{guerreiro-etal-2024-xcomet} provide transparent MT evaluation through fine-grained error detection, while GEMBA-MQM \citep{kocmi-federmann-2023-gemba} leverages high-capacity LLMs to identify error spans without human references. To enhance interpretability, xTower \citep{guerreiro-etal-2024-xtower} provides free-text explanations for translation errors. In the summarization domain, G-Eval \citep{liu-etal-2023-geval} utilizes chain-of-thought prompting to evaluate summaries across multiple dimensions like coherence and relevance, achieving high human alignment. However, these methods often act as "black-box" scorers or require task-specific error taxonomies.

More recent reference-free approaches move beyond QA: \citet{wu-etal-2023-holistic} combine semantic, fluency, and adequacy signals into a holistic scorer, while \citet{moosa-etal-2024-mtranker} propose MT-Ranker, a pairwise ranking model that evaluates translations by direct comparison, thus achieving strong performance without references or absolute scores. Our Cross-Examination Framework (CEF) builds on this line of work while addressing its key limitations. Unlike reference-based methods, CEF is fully reference-free. Distinct from multi-step pipelines like FEQA and QAGS and unlike ranking-only approaches like MT-Ranker, it integrates question generation and answering into a single, robust process to reduce error propagation. By generating verifiable questions from both the source and the translation, CEF provides a more principled and scalable approach to directly evaluating factual consistency and content coverage.

\begin{table*}[htbp!]
\centering
\begin{adjustbox}{width=\textwidth,center}
\begin{tabular}{l|cccc|cc|cc}
\toprule
\multirow{3}{*}{Model} & \multicolumn{4}{c|}{\textbf{Translation (NTREX)}} & \multicolumn{2}{c|}{\textbf{Summarization}} & \multicolumn{2}{c}{\textbf{Note Generation}} \\
& \multicolumn{4}{c|}{\textbf{(Source Questioning)}} & \multicolumn{2}{c|}{\textbf{(CNN/DailyMail)}} & \multicolumn{2}{c}{\textbf{(ACI-Bench)}} \\
\cmidrule(lr){2-5} \cmidrule(lr){6-7} \cmidrule(lr){8-9}
& en-fr & en-es & en-ar & en-jp & Source & Summary & Source & Note \\
\midrule
\multicolumn{9}{c}{\textbf{Disagreement Rates (ADR) (with rank)}} \\
\midrule
Llama-3.3-70B   & 0.89 (1) & 1.12 (4) & 2.36 (2) & 2.70 (3) & 0.76 (2) & 0.87 (3) & 1.92 (4) & 0.53 (4) \\
Qwen3-235B      & 1.46 (4) & 1.28 (5) & 3.50 (5) & 3.54 (5) & 1.26 (5) & 2.30 (5) & 3.69 (5) & 1.06 (5) \\
DeepSeek-V3     & 1.28 (3) & 1.00 (2) & 2.26 (1) & 1.75 (1) & 0.70 (1) & 0.87 (3) & 1.36 (2) & 0.31 (1) \\
GPT-OSS-120B    & 1.08 (2) & 1.10 (3) & 2.52 (4) & 2.70 (3) & 0.93 (4) & 0.74 (1) & 0.97 (1) & 0.33 (2) \\
Gemma-3-27B     & 1.65 (5) & 0.87 (1) & 2.46 (3) & 1.91 (2) & 0.82 (3) & 0.77 (2) & 1.45 (3) & 0.49 (3) \\
\midrule
\multicolumn{9}{c}{\textbf{Deviation Scores (ADS) (with rank)}} \\
\midrule
Llama-3.3-70B   & 0.81 (5) & 0.96 (4) & 1.80 (4) & 1.96 (5) & 0.09 (1) & 0.22 (5) & 0.66 (3) & 0.55 (5) \\
Qwen3-235B      & 0.40 (3) & 0.27 (3) & 0.45 (3) & 0.41 (3) & 0.10 (3) & 0.17 (3) & 1.09 (5) & 0.19 (3) \\
DeepSeek-V3     & 0.20 (1) & 0.10 (1) & 0.39 (2) & 0.15 (1) & 0.11 (4) & 0.15 (2) & 0.36 (1) & 0.17 (1) \\
GPT-OSS-120B    & 0.67 (4) & 1.07 (5) & 1.97 (5) & 1.57 (4) & 0.15 (5) & 0.18 (4) & 0.75 (4) & 0.17 (1) \\
Gemma-3-27B     & 0.35 (2) & 0.14 (2) & 0.22 (1) & 0.29 (2) & 0.09 (1) & 0.13 (1)& 0.52 (2) & 0.21 (4) \\
\bottomrule
\end{tabular}
\end{adjustbox}
\caption{Task-agnostic reliability analysis showing Disagreement Rates (ADR) and Deviation Scores (ADS) across Translation (NTREX), Summarization (CNN/DailyMail), and Clinical Note Generation (ACI-Bench). DeepSeek-V3 consistently emerges as the most stable judge across diverse text-to-text tasks.}
\label{tab:task_agnostic_reliability}
\end{table*}

\section{Methodology} \label{methodology}
The Cross-Examination Framework was originally introduced for evaluating clinical summarization \cite{kanithi2024medic}. In our work, we adapt it for machine translation, summarization and note-generation. The framework operates in three main stages:

\begin{enumerate}
    \item \textbf{Question generation:} Closed-ended questions are generated independently from the source document $ D $ and the generated candidate $ T $ by prompting an LLM. Specifically, we define $Q_D = {q_i \mid i = 1, \ldots, N}$ as the set of $N$ questions generated from $D$, and $Q_T = {q_j \mid j = 1, \ldots, N'}$ as the set of $N'$ questions generated from $T$.
    % Specifically, $ N $ questions $ q_i \in Q_D $ are generated from the original document while $ N' $ questions $ q_j \in Q_T $ are generated from the generated candidate.
    All questions are designed such that the correct answer is "YES", ensuring that they are verifiable and grounded in the content of the originating document. 
    
    \item \textbf{Cross-Examination: } In this
    step, questions generated from the document $ Q_D $ are presented to the generated candidate $ T $ yielding a set of answers $ \hat{A}_{Q_D|T} $ where each $ \hat{a}_i \in \{"YES", "NO", "IDK"\}$. Similarly, $ \hat{A}_{Q_T|D} $ is generated by presenting the questions generated from the generated text to the document.
    
    \item \textbf{Score Computation: } In the final step, the predicted answers from the previous step are compared against the ground-truth "YES" labels to derive the following scores:

    \textit{Coverage:} Measures how comprehensively the generation $ T $ covers the content of the original document $ D $. The score is calculated as $100 - \%(\hat{A}_{Q_D|T} == "IDK")$, where a higher score indicates that the generated text retains more of the original document's details.
    
    \textit{Consistency:} This score is calculated as $ 100 - \%(\hat{A}_{Q_T|D} == "IDK")$, which measures the degree of hallucination in the generated content.
    
    \textit{Conformity:} This score calculated as $ 100 - \%(\hat{A}_{Q_D|T} == "NO")$, measures if the generation introduces contradiction with respect to the original document.

    \textit{Conciseness:} Applied primarily to summarization tasks, this score reflects the brevity of the output and is computed by the reduction in word count from the original document to the candidate.
    % \begin{itemize}
    %     \item \textit{Coverage:} Measures how comprehensively the translated text $ T $ covers the content of the original document $ D $. The score is calculated as $100 - \%(\hat{A}_{Q_D|T} == "IDK")$, where a higher score indicates that the translation retains more of the original document's details.
    %     \item \textit{Consistency:} This score is calculated as $ 100 - \%(\hat{A}_{Q_T|D} == "IDK")$, which measures the degree of hallucination in the translated text.
    %     \item \textit{Conformity:} This score calculated as $ 100 - \%(\hat{A}_{Q_D|T} == "NO")$, measures if the translated text introduces contradiction with respect to the original document.
    % \end{itemize}

\end{enumerate}

One of the key strengths of CEF is its interpretability. Conflicting questions where predictions diverge when grounded on the source document serve as diagnostic tools, enabling fine-grained analysis of errors such as omissions, additions, or semantic inversions. Prompts for question generation and cross-examination are provided in Appendix \ref{app:prompts}. CEF operates primarily in two modes: \textbf{Reference-Free Mode}, where questions are generated from the source document and candidate assess fidelity without a gold standard, and \textbf{With-Reference Mode}, where questions are generated from the candidate and a reference candidate to evaluate their specific alignment.

\subsection{CEF Reliability} \label{meth:robustness}

% The question-generation and answering components of CEF are central to its effectiveness. Therefore, it is essential to verify that these components produce stable and robust outputs, particularly for a specific language or domain. Additionally, identifying the most suitable model to serve as the judge is critical for reliable results.

% Each model $M_i$, where $i \in \{1, 2, 3, 4, 5\}$, independently generates a set of questions $Q_i$ from the same document $D$. For each question $q \in Q_i$, every model $M_j$ with $j \in \{1, 2, 3, 4, 5\}$ provides an answer $A_j(q)$. We define a disagreement between two answers $A_i(q)$ and $A_j(q)$ as:

% {\small
% \begin{equation*}
% \text{Disagreement}(A_i(q), A_j(q)) =
% \begin{cases}
% 1 & \text{if } A_i(q) \neq A_j(q) \\
% 0 & \text{otherwise}
% \end{cases}
% \end{equation*}
% }

The reliability of CEF depends on the stability of its question-generation and answering components. To assess this, we evaluate whether questions elicit consistent answers across different models and whether candidate judges agree with the consensus.  

Let $\mathcal{M} = \{M_1, \ldots, M_K\}$ denote a set of $K$ models. Each model $M_i$ generates a set of questions $Q_i = \{q_1, \ldots, q_{N_i}\}$ from a document $D$. For each question $q \in Q_i$, every model $M_m \in \mathcal{M}$ provides an answer $A_m(q) \in \{\text{YES}, \text{NO}, \text{IDK}\}$ with document $D$ provided as context.  

We define pairwise disagreement ($\delta$) between two models $M_m$ and $M_n$ on question $q$ as:  

% {
% \[
% \delta(A_m(q), A_n(q)) =
% \begin{cases}
% 1 & \text{if } A_m(q) \neq A_n(q), \\
% 0 & \text{otherwise}.
% \end{cases}
% \]
% }  
{
\[
\delta_{mn}(q) =
\begin{cases}
1 & \text{if } A_m(q) \neq A_n(q), \\
0 & \text{otherwise}.
\end{cases}
\]
}  

% We calculate the average disagreement across the judges and answer deviation to quantify the reliability of the question generation and answering components:
Using this, we introduce two measures: 

% \begin{itemize}
    % \item \textbf{Average Disagreement Rate (ADR):} For a given model $ M_i $, we compute its average disagreement across all questions it generates. Specifically, for each $ q \in Q_i $, we calculate the fraction of other models that provide a different answer, and then average over all questions in $ Q_i $. A lower $ADR(M_i)$ indicates that $M_i$'s questions elicit more consistent answers across models, suggesting higher reliability in question formulation.
    % {\small
    % \begin{equation*}
    % \begin{split}
    % \text{AvgDisagreementRate}(M_i) = \frac{1}{|Q_i|} \sum_{q \in Q_i} \bigg( \frac{1}{4} \sum_{\substack{j=1 \\ j \neq i}}^{5} \\
    % \text{Disagreement}(A_i(q), A_j(q)) \bigg)
    % \end{split}
    % \end{equation*}
    % }
    \textit{Average Disagreement Rate (ADR):} For model $M_i$,  
    % \[
    % \begin{split}
    % \text{ADR}(M_i) &= \frac{1}{|Q_i|} \sum_{q \in Q_i} \frac{1}{K-1} \\
    % &\quad \sum_{\substack{m=1 \\ m \neq i}}^K \text{Disagreement}(A_m(q), A_i(q))
    % \end{split}
    % \]
    {
    \[
    \begin{split}
    \text{ADR}(M_i) &= \frac{1}{|Q_i|} \sum_{q \in Q_i} \frac{1}{K-1} \sum_{\substack{m=1 \\ m \neq i}}^K \delta_{mi}(q)
    \end{split}
    \]
    }
    ADR quantifies how often questions from $M_i$ yield inconsistent answers across other models. Lower values indicate more stable question generation.

    % \item \textbf{Answer Deviation Score  (ADS):} The Answer Deviation Score for a model \( M_i \) measures how often the model's answers differ from the most common (consensus) answer across all models. It is calculated by taking the average disagreement between the model's answer and the majority answer for each question, over the entire set of questions. A higher score indicates that the model deviates more frequently from the consensus.
    % This score for a particular model $M_i$ can be defined as:
    
    % {\small
    % \begin{equation*}
    % \begin{split}
    % \text{AnswerDeviationScore}(M_i) = \frac{1}{|Q_{all}|} \sum_{q \in Q_{all}} \\ \text{Disagreement}(A_i(q), A_{maj}(q))
    % \end{split}
    % \end{equation*}
    % }
    % where $A_{maj}(q)$ denotes the the most frequent answer to $q$ and $Q_{all}$ denotes the union of all questions from all models. This score quantifies how frequently $M_i$ deviates from the consensus answer.
    \textit{Answer Deviation Score (ADS):} For model $M_i$, let $A^*(q)$ denote the majority answer across $\mathcal{M}$. Then,  
    \[
    ADS(M_i) = \frac{1}{|Q|} \sum_{q \in Q} \mathbf{1}[A_i(q) \neq A^*(q)],
    \]  
    where $Q = \bigcup_{i=1}^K Q_i$. ADS measures how often $M_i$’s answers deviate from the consensus across all questions.  

% \end{itemize}

% We can argue that the models with lower ADR and ADS can be considered as good judges. On the other hand, if the scores are higher across different models it suggests that the questions and answers generated are inconsistent for that particular setting therefore making CEF unreliable in that scenario. 

Models with low ADR and ADS are preferred as judges. High values suggest instability in question generation or answer prediction, indicating unreliable evaluation under those conditions.

\section{Experiments} \label{experiments}

\begin{table*}[htbp]
\centering
\begin{tabular}{l | *{3}{c} | *{3}{c}}
\toprule
\multirow{2}{*}{Num Questions} & 
\multicolumn{3}{c}{Original} & 
\multicolumn{3}{c}{Qwen3-1.7B} \\
\cmidrule(lr){2-4} \cmidrule(lr){5-7}
& Conformity & Consistency & Coverage & Conformity & Consistency & Coverage \\
\midrule
3  & 15.96 & 46.35 & 32.65 & 40.22 & 145.08 & 55.48 \\
5  & 10.54 & 19.22 & 25.97 & 41.07 & 120.02 & 69.30 \\
10 & 4.02 & 9.78 &  9.94 & 11.78 &  15.18 & 20.73 \\
20 & 4.35 & 9.23 & 8.32 & 9.45 & 15.82 & 20.25 \\
\bottomrule
\end{tabular}
\caption{Variance of CEF scores for en-jp across different question counts. Low $N$ (3–5) leads to high instability, especially for noisy Qwen3-1.7B translations, while $N=10$ provides similar stability to costlier $N=20$.}
\label{tab:variance_comparison}
\end{table*}

%% How reliable is the reference free mode. Evaluate across different language pairs and models. Report the best judge based on the reliability metric. Do try some bad models and some low-resource language pairs. Comment on the reliability of the different language pairs.
\subsection{Datasets and models}

We evaluate the Cross-Examination Framework across three distinct tasks. For machine translation, we use the NTREX dataset \citep{federmann-etal-2022-ntrex, barrault-etal-2019-findings}, comprising 123 English news documents translated into 128 languages. We focus our analysis on four target languages, French, Spanish, Arabic and Japanese, to cover a range of scripts and morphological complexity. For abstractive summarization, we used the CNN/DailyMail test subset (11.5k samples) \citep{see-etal-2017-get} and for clinical note generation, we used the ACI-Bench dataset (120 samples) \citep{yim2023acibenchnovelambientclinical}.

For the robustness analysis, we benchmark five Large Languages Models as candidate judges: \texttt{Llama-3.3-70B-Instruct} \cite{grattafiori2024llama3herdmodels}, \texttt{Qwen3-235B-A22B-Instruct} \cite{qwen3technicalreport}, \texttt{GPT-OSS-120B} \cite{openai2025gptoss120bgptoss20bmodel}, \texttt{DeepSeek-V3} \cite{deepseekai2024deepseekv3technicalreport}, and \texttt{Gemma-3-27B-IT} \cite{gemmateam2025gemma3technicalreport}. Finally, we validate CEF against human judgment by comparing mismatching questions with exper-annotated error spans from WMT'25 (Japanese subset) \citep{kocmi-etal-2025-findings} and factuality labels from the FRANK benchmark (filtered for CNN-DM) \citep{pagnoni-etal-2021-understanding}.

\subsection{Robustness of the CEF Framework} \label{exp:robustness}

The reliability of the CEF hinges on its judge model. To select the most robust model, we conducted a verification analysis across the three tasks. Each candidate judge generated $N=10$ questions using deterministic decoding ($temperature = 0$) to ensure results were verifiable and grounded strictly in document content. We then calculated inter-model agreement using the Disagreement Rate (ADR) and Answer Deviation Score (ADS) to identify the model most aligned with the multi-model consensus.

As shown in Table \ref{tab:task_agnostic_reliability}, \texttt{DeepSeek-V3} consistently achieves the lowest disagreement and deviation scores across nearly all language pairs and task types. In the clinical note generation task, it demonstrates superior stability with an ADR of 0.31 and an ADS of 0.17 when querying generated notes. While \texttt{Llama-3.3-70B} and \texttt{GPT-OSS-120B} show competitive stability in specific metrics, such as the 0.09 ADS for summarization source questioning, they exhibit significantly higher variability in cross-lingual settings (Arabic and Japanese) and clinical note generation. \texttt{Qwen3-235B} consistently yielded the highest disagreement rates in most of the tasks. Given its superior consistency across both multilingual and multi-task settings, we selected \texttt{DeepSeek-V3} as the judge for all subsequent experiments to ensure a stable, task-agnostic evaluation baseline.

\subsection{Evaluating Text-to-Text Quality}
\label{exp:main}

% \begin{figure*}[htbp]
\begin{figure*}[t!]
    \centering
    \includegraphics[width=\textwidth]{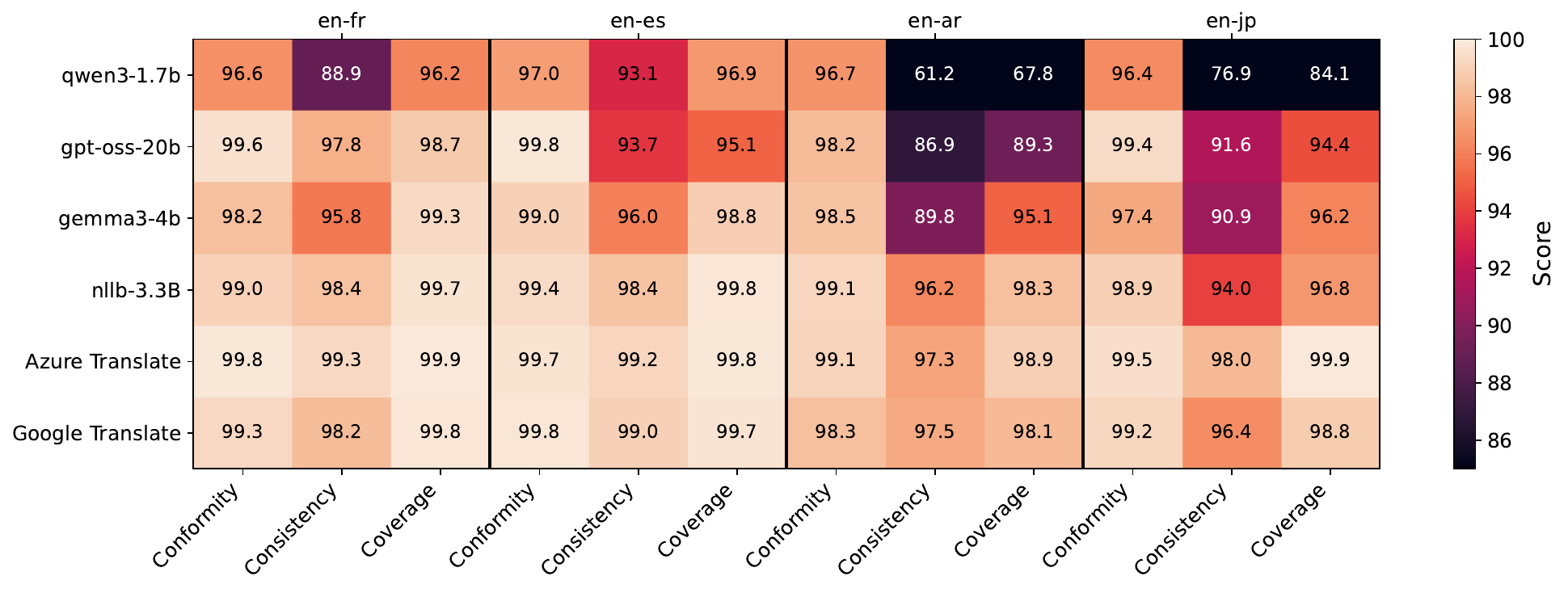} \\[0.8em]
    \includegraphics[width=\textwidth]{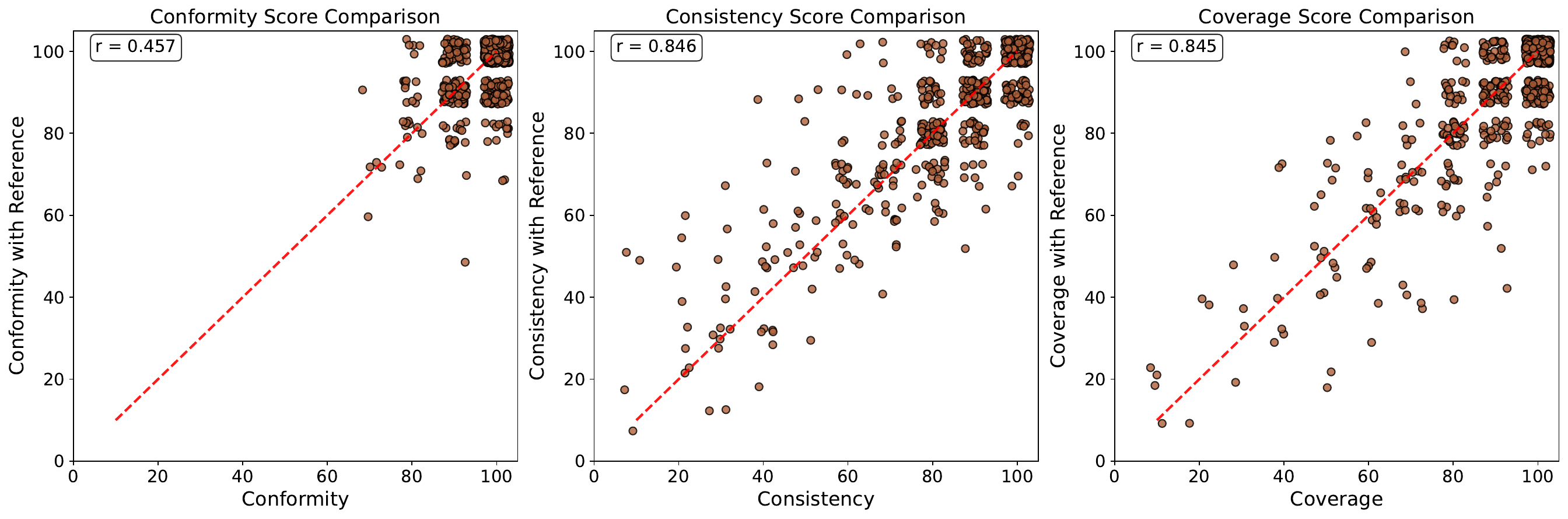}
    \caption{\textbf{(Top)} Reference-free CEF evaluation using DeepSeek-V3 as judge across four languages, where NLLB-3.3B along with commercial systems (Google, Azure) show strong performance while performance decreases for Arabic and Japanese. \textbf{(Bottom)} Correlation between reference-free and with-reference CEF scores for Consistency and Coverage ($r=0.846, 0.845$), validating CEF’s reliability without gold references, while Conformity shows weaker correlation ($r=0.457$) due to lower variability.}
    \label{fig:cef_combined}
\end{figure*}

\begin{figure}[t!]
    \centering
    \includegraphics[width=0.48\textwidth]{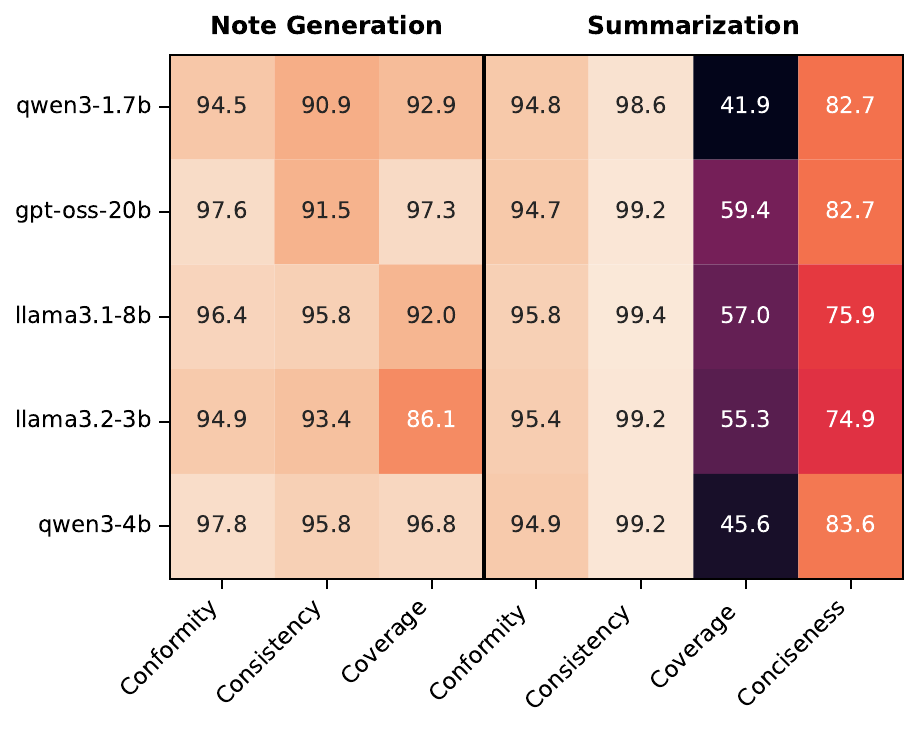} 
    \caption{Reference-free CEF performance across Clinical Note Generation and Abstractive Summarization tasks}
    \label{fig:cef_note_summ}
\end{figure}

Using our selected judge, \texttt{DeepSeek-V3}, we first evaluate translations in \textbf{reference-free} mode. We assess translations across four target languages (French, Spanish, Arabic, Japanese) generated by: three diverse LLMs (\texttt{google/gemma-3-4b-it} \cite{gemmateam2025gemma3technicalreport}, \texttt{Qwen/Qwen3-1.7B} \cite{qwen3technicalreport}, \texttt{openai/gpt-oss-20b} \cite{openai2025gptoss120bgptoss20bmodel}), a specialized translation model \texttt{facebook/nllb-200-3.3B \cite{nllbteam2022languageleftbehindscaling}} and commercial API services (Azure Translate, Google Translate). As shown in Figure \ref{fig:cef_combined}, specialized models like \texttt{NLLB-3.3B} and commercial APIs consistently outperform general-purpose LLMs. This performance gap widens for morphologically complex languages like Arabic and Japanese, where all models exhibit lower consistency and coverage, highlighting CEF's sensitivity to language-specific challenges. Beyond aggregate scores, CEF offers interpretability by pinpointing errors through specific question-answer mismatches (see Appendix \ref{app:cef_example}).

In clinical note generation (Figure \ref{fig:cef_note_summ}), the results show that larger models like gpt-oss-20b achieve the highest Coverage (97.33), indicating they are highly effective at retaining specific details from patient-physician dialogues. However, models like qwen3-4b show stronger Consistency (95.83), suggesting a better ability to avoid introducing extrinsic information that was not present in the original conversation.

In the summarization task, CEF uncovers a clear relationship between information density and content retention. Models that produce more concise summaries, such as qwen3-1.7b (Conciseness: 82.7), often do so at the cost of lower Coverage (41.9). This indicates that the models are prioritizing brevity by omitting certain semantic details. Across all summarization models, Consistency scores remain exceptionally high (all above 98.6), while Coverage scores range from 41.89 to 59.36.

To validate this reference-free approach, we compare its outputs to a \textbf{with-reference mode}, where CEF evaluates a system's output against a human reference for the translation task. Figure \ref{fig:cef_combined} reveals strong positive correlations between the two modes for Coverage ($r=0.845$) and Consistency ($r=0.846$), confirming that our reference-free assessment reliably measures content fidelity without gold references. The lower correlation for Conformity ($r=0.457$) is expected; as contradictions are rare, scores cluster near 100\%, which limits variance and naturally lowers the Pearson correlation.

% \begin{figure}[htbp]
\begin{figure}[t!]
    \centering
    \includegraphics[width=0.48\textwidth]{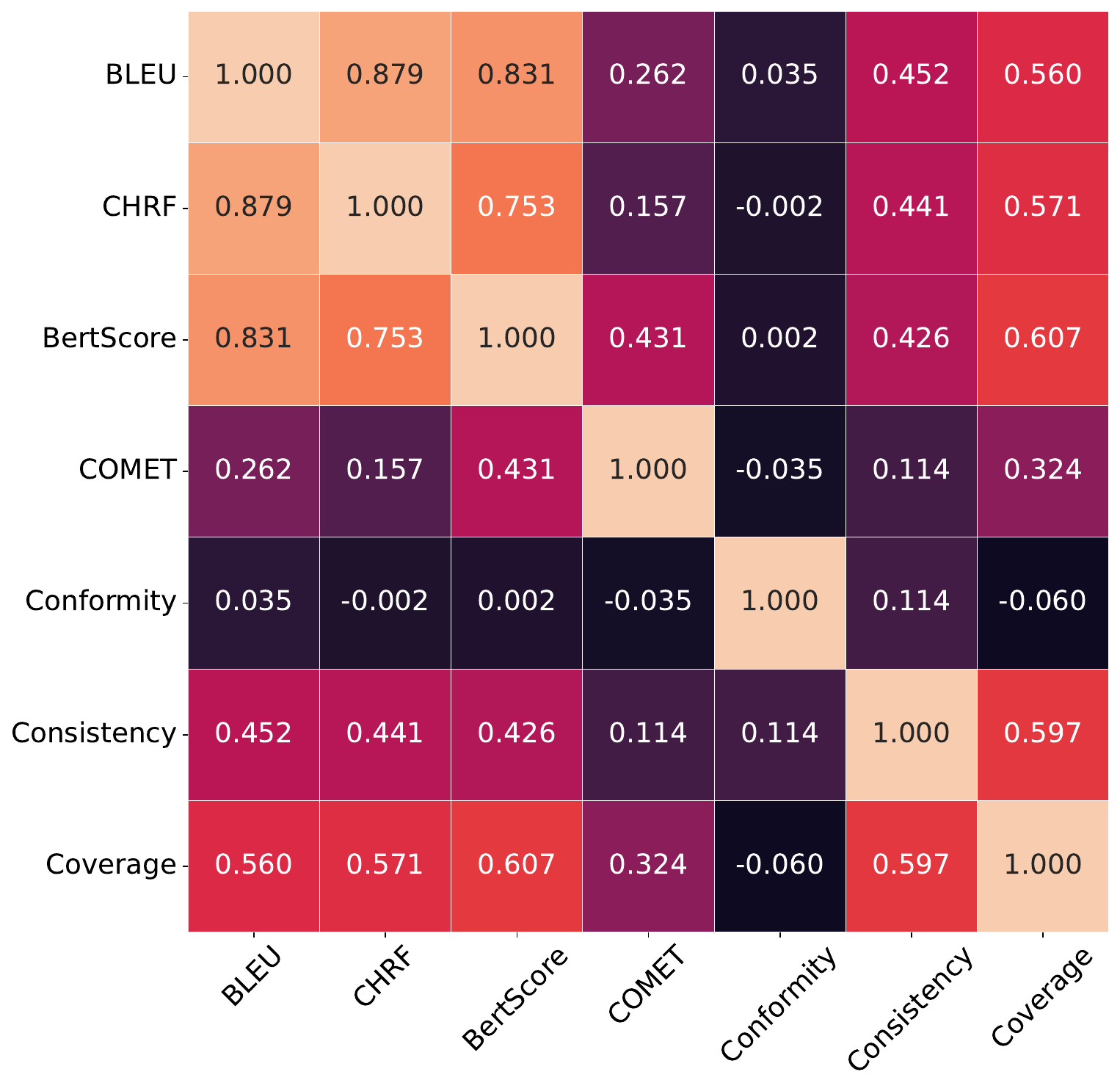}
    \caption{CEF metrics (Coverage, Consistency) show stronger correlation with semantic metrics (BertScore), than surface metrics (BLEU/chrF) or Conformity, validating CEF’s unique focus on deep semantic fidelity beyond lexical overlap.}
    \label{fig:cef_trad_correlation_heatmap}
\end{figure}

Finally, we position CEF against traditional metrics by correlating its scores with BLEU, chrF, BertScore and COMET. The heatmap in Figure \ref{fig:cef_trad_correlation_heatmap} shows that CEF provides complementary signals. Surface-level metrics like BLEU correlate only moderately with CEF's semantic scores. BertScore, being semantic, shows a stronger alignment. This analysis validates CEF's unique contribution: it excels at measuring deep information fidelity, particularly factual consistency and content completeness, where traditional metrics are known to be less effective.

\subsection{Human validation}
% Table: WMT'25 Human Alignment (Translation)
\begin{table}[ht]
\centering
\small
\begin{tabular}{lccc}
\toprule
\textbf{Error Type} & \textbf{Count} & \textbf{CEF Align.} & \textbf{Align. Rate} \\ \midrule
Semantic & 852 & 668 & \textbf{78.40\%} \\
Non-semantic & 2,316 & 540 & \textbf{23.32\%} \\ \midrule
Overall & 3,168 & 1,208 & \textbf{38.13\%} \\ \bottomrule
\end{tabular}
\caption{Alignment between human expert annotations from WMT'25 (Japanese subset) and CEF mismatching questions. CEF shows a significantly higher alignment rate for errors that directly affect semantic fidelity.}
\label{tab:wmt_alignment}
\end{table}

% Table: FRANK Human Alignment (Summarization)
\begin{table}[ht]
\centering
\small
\begin{tabular}{lccc}
\toprule
\textbf{Error Type} & \textbf{Caught} & \textbf{Total} & \textbf{Percentage} \\ \midrule
Entity & 144 & 167 & 86.2\% \\
Coreference & 29 & 85 & 34.1\% \\
Grammatical & 17 & 64 & 26.6\% \\
Out of Article & 33 & 38 & 86.8\% \\
Relation & 35 & 38 & 92.1\% \\
Circumstance & 28 & 33 & 84.8\% \\
Discourse Link & 7 & 17 & 41.2\% \\
Other & 2 & 7 & 28.6\% \\ \midrule
\textbf{OVERALL} & \textbf{298} & \textbf{449} & \textbf{65.7\%} \\ \bottomrule
\end{tabular}
\caption{CEF alignment with factual error categories in the FRANK benchmark for abstractive summarization. The framework excels at identifying relational and entity-based semantic distortions.}
\label{tab:frank_alignment}
\end{table}

To validate the diagnostic utility of the Cross-Examination Framework, we evaluate the correspondence between CEF mismatching questions and human-annotated errors across translation and summarization tasks. This analysis investigates whether the framework's semantic cross-examination successfully identifies the specific error spans flagged by expert annotators.

 We use the Japanese subset of the WMT’25 dataset, which provides fine-grained human expert annotations of translation errors. We focus on the English-Japanese direction to maintain consistency with our core experiments. We use DeepSeek-V3 to identify a correspondence between a CEF mismatch and a human-annotated error that points to the same underlying issue or demonstrates a clear conceptual overlap and further classify each human-annotated error to check if the error affects semantic fidelity. As shown in Table~\ref{tab:wmt_alignment}, the CEF aligns with semantic errors (78,49\%) at a rate  3.4 times higher than with non-semantic error (23.32\%). Although most human annotations in the dataset focus on grammar or style, the CEF's higher sensitivity to meaning-altering errors suggests it effectively targets failures in information fidelity.
 
 % As shown in Table~\ref{tab:wmt_alignment}, CEF demonstrates exceptional alignment with semantic errors (78.40\%), which is more than 3.4 times higher than its alignment with non-semantic errors (23.32\%). While the majority of human annotations pertain to non-semantic issues such as grammatical or stylistic errors, CEF remains highly targeted toward the errors that directly affect meaning, which validates that CEF captures the specific failure modes pertaining to information fidelity.

For the summarization task, we evaluate CEF against the FRANK benchmark, which categorizes factual errors in model-generated summaries of the CNN/DailyMail dataset. The results, detailed in Table~\ref{tab:frank_alignment}, reveal that CEF is particularly effective at identifying fact-based semantic distortions. The framework achieves its highest alignment rates in categories involving relational and entity-based facts, such as \textit{Relation} (92.1\%), \textit{Out of Article} (86.8\%), and \textit{Entity} (86.2\%) errors.

These high rates suggest that the CEF is highly sensitive to "hard" hallucinations, such as the introduction of external information or the misrepresentation of predicate-argument structures between entities. In contrast, the CEF shows lower alignment with \textit{Grammatical} (26.6\%) and \textit{Coreference} (34.1\%) errors which shows CEF prioritizes global semantic grounding over surface-level linguistic fluency.

\subsection{Ablation on the number of questions}
\begin{figure*}[htbp]
    \centering
    \includegraphics[width=0.48\textwidth]{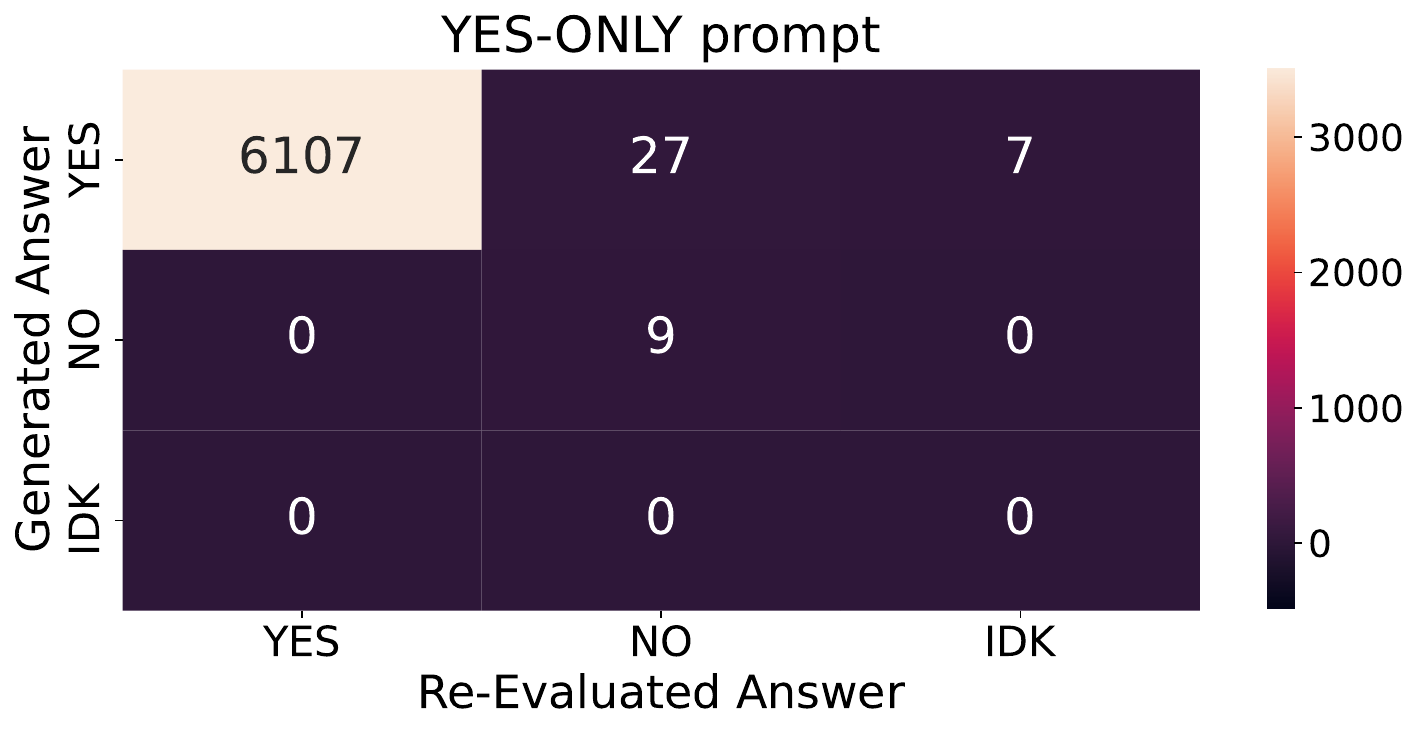}\hfill
    \includegraphics[width=0.48\textwidth]{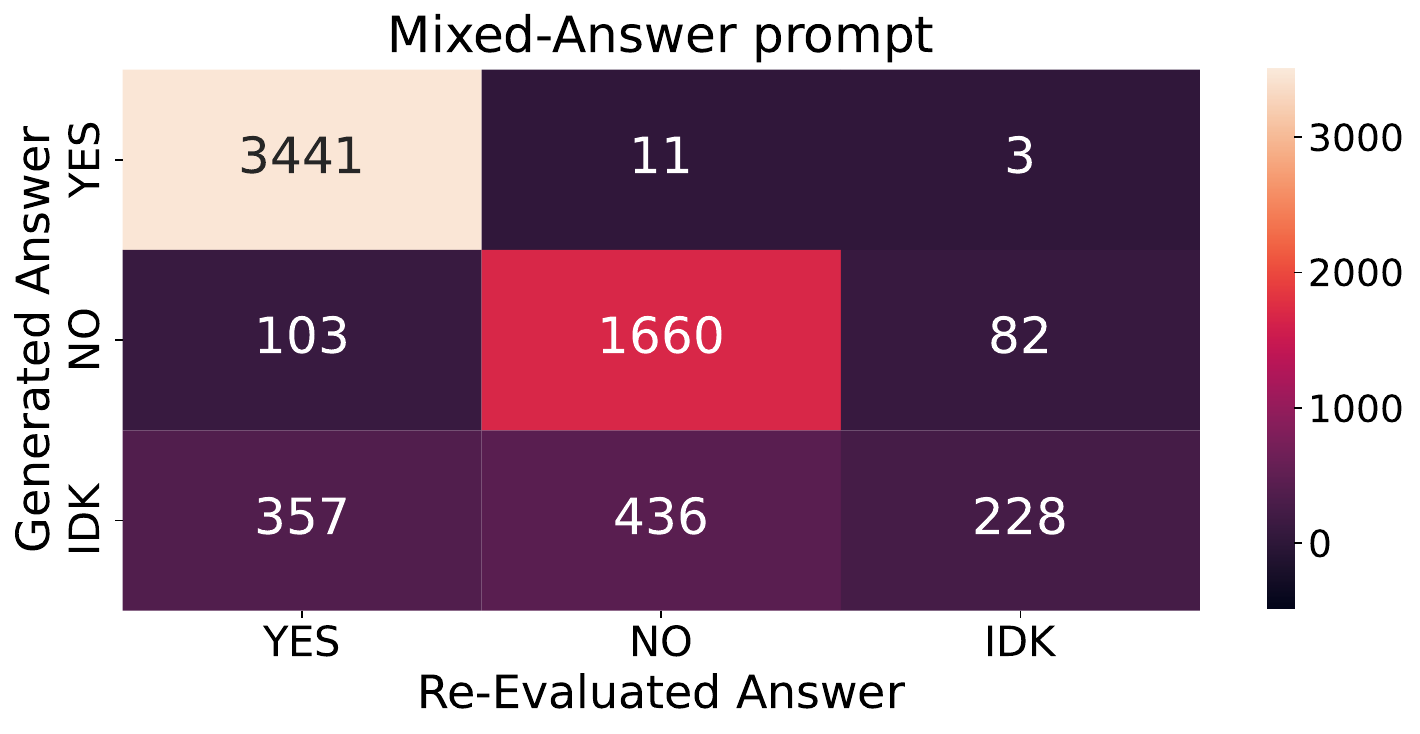}
    \caption{Comparison of generated versus re-evaluated answers for two prompting strategies: a Yes-only prompt (producing only “YES” answers) and a Mixed-answer prompt (allowing “YES”, “NO”, or “IDK”). Each confusion matrix illustrates how the initially generated answers align with re-evaluated answers when the same document is given as a context during re-evaluation.
    \textbf{(left)} YES-Only Prompt demonstrates high stability, while 
    \textbf{(right)} Mixed-Answer Prompt shows significant instability for "NO" and "IDK" answers.}
    \label{fig:confusion_comparison}
\end{figure*}

A critical parameter in the Cross-Examination Framework is the number of questions generated per document ($N$), which directly influences score stability and computational cost. To determine the optimal $N$ , we conducted an ablation study measuring the standard deviation of CEF scores across 10 independent runs for $N \in \{3,5,10,20\}$. Results for Japanese translations (en-jp) on the NTREX dataset, representing a morphologically complex distant language pair, are summarized in Table~\ref{tab:variance_comparison}—comparing human-translated references ("Original") against noisy translations from \texttt{Qwen3-1.7B}.

The ablation study reveals critical patterns in how question quantity affects evaluation stability. At low question counts (N=3-5), all CEF scores exhibit severe instability, with variance reaching 46.35 for Consistency in human translations and escalating to 145.08 for noisy Qwen3-1.7B outputs at N=3. This confirms that sparse question sampling fails to robustly capture document semantics. While increasing N to 20 yields marginal stability gains—such as a mere 1.62-point Coverage improvement for human translations—doubling the computational cost from N=10 proves unjustified, as N=10 already achieves near-optimal reliability by reducing Consistency variance by 80\% compared to N=5 for reference translations. Notably, translation quality directly impacts score stability: noisy outputs consistently show 2–3× higher variance than human references (e.g., Conformity SD of 40.22 vs. 15.96 at N=3 ).

These findings establish N=10 as the optimal balance between statistical reliability and computational efficiency for cross-lingual evaluation. Since all translations in the NTREX dataset are derived from the same set of English source documents, the underlying semantic content and document length distribution are consistent across languages. Consequently, we adopt N=10 for all subsequent experiments, ensuring robust evaluation across diverse language pairs. Additional ablation results for other language pairs are provided in \autoref{tab:variance_comparison_fr_es}.

\subsection{Ablation Study: Impact of "YES-Only" Question Constraints}
\label{exp:ablation_yes}

% Comparison of Generated vs. Re-evaluated answers across prompt types. 
To assess the efficacy of enforcing "YES-only" answers during question generation, we compared two prompting strategies: (1) a constrained prompt generating exclusively "YES" questions and (2) an unconstrained prompt allowing answers from $\{"YES","NO","IDK"\}$ on the NTREX dataset. For both conditions, we generated $N = 10$ questions per document and re-evaluated their answers using the same document as context. Confusion matrices in Figure ~\ref{fig:confusion_comparison} compare the generated answer (rows) with the re-evaluated answer (columns).

The original "YES-only" prompt demonstrates exceptional stability: 99.4\% of 6141 "YES" questions retain their label upon re-evaluation, with minimal shifts to "NO" (27 instances) or "IDK" (7 instances). This confirms that constraining questions to "YES" ensures strong factual grounding, as the LLM reliably generates questions directly verifiable from document content.

In contrast, the modified prompt reveals critical limitations in generating non-"YES" questions:

\begin{itemize}
    \item "NO" Questions: 5.6\% of 1,845 "NO" questions revert to "YES", indicating the LLM struggles to formulate verifiable negative claims. Many generated "NO" questions inadvertently reference unstated implications rather than explicit contradictions.
    \item "IDK" Questions: Only 22.3\% of 1021 "IDK" questions remain unchanged, with 357 shifting to "YES" and 436 to "NO". Crucially, this instability stems from the LLM's inability to generate genuinely unanswerable questions when conditioned on the document. Rather than creating questions outside the document's scope, the model produces questions plausibly related to the content but ambiguously phrased, which the re-evaluation step correctly resolves as "YES" or "NO" using contextual inference.
\end{itemize}

These findings validate the necessity of the "YES-only" constraint for reliable evaluation. By restricting questions to verifiable facts, we eliminate the LLM's tendency to generate ambiguous or unanswerable questions that undermine score reliability. While mixed-answer prompting appears theoretically appealing, its practical implementation fails due to the model's contextual grounding bias—making "YES-only" not merely a simplification, but a requirement for robust cross-examination.

\section{Conclusion}

In this work, we introduced the Cross-Examination Framework (CEF) as a novel, task-agnostic approach for assessing information fidelity in text-to-text generation. Unlike traditional metrics that focus primarily on surface similarity, CEF assesses semantic fidelity through three interpretable dimensions: Coverage, Consistency, and Conformity. Across a wide range of text-to-text tasks, CEF demonstrated strong alignment with semantic similarity metrics like BERTScore while exposing critical errors such as omissions, hallucinations, and contradictions that conventional methods frequently miss.

Our contributions include a systematic robustness analysis of judge models, where DeepSeek-V3 proved to be the most stable and reliable choice. We also identified an optimal configuration of ten questions per document, balancing statistical stability with computational efficiency. Furthermore, we validated the framework against human judgment, showing that CEF mismatches correlate strongly with expert-annotated semantic errors. While primarily reference-free, CEF's reliability is further supported by the strong correlation observed between its reference-free and with-reference modes. These findings establish CEF as a valuable complementary metric that enriches evaluation by providing deeper semantic insights rather than serving as a replacement for existing metrics.

\section{Limitations}
Despite its advantages, CEF has several limitations. First, its reliability is closely tied to the choice of judge model, which introduces potential bias due to the training data and alignment of that model. In low-resource languages (Appendix \ref{exp:round_trip}), unstable or incoherent question generation remains a bottleneck, and the multi-step pipeline is more computationally demanding than traditional metrics and LLM-based metrics (Appendix \ref{app:token_analysis}).

To address these issues, future work should explore experiments with multilingual small or distilled LLMs as judges to reduce cost and bias. Improving multilingual question generation will also be critical to expanding CEF’s applicability across underrepresented languages. Furthermore, incorporating multi-hop reasoning questions could enhance the framework’s ability to capture complex factual relations. Finally, integrating CEF with emerging evaluation paradigms such as automatic factuality detection and preference modeling could establish a unified framework for holistic assessment of information fidelity in natural language generation.

\bibliography{custom}

\newpage
% \onecolumn
\appendix
% \onecolumn
\section{Appendix}

\subsection{Experimental settings}
For the experiments in Section~\ref{exp:robustness}, all judge models generated questions and answers with temperature set to 0 to minimize sampling variance. Inference was performed using vLLM~\cite{kwon2023efficient} on a node equipped with 8 NVIDIA H200 GPUs. The same vLLM backend was used to generate translations from all LLMs in other experiments. For Sections~\ref{exp:main} and~\ref{exp:round_trip}, we employed Hugging Face Transformers~\cite{wolf2019huggingface} with the \texttt{pipeline} API for inference on NLLB-200. Google Translate evaluations utilized the Advanced Cloud Translation API, while Azure Translate was accessed via the \texttt{TextTranslationClient} from the Azure AI Translation Text SDK. For all traditional metrics, we used the pipelines from the Hugging Face \texttt{evaluate} library: \texttt{sacrebleu} \cite{post-2018-call} for BLEU scores, \texttt{chrf} for chrF scores, \texttt{bertscore} for BERTScore, and \texttt{comet} for COMET scores. For BLEU, we used the \texttt{flores101} tokenizer \cite{goyal-etal-2022-flores}, as it supports all languages in our evaluation. For BERTScore, we used the \texttt{bert-base-multilingual-cased} model \cite{DBLP:journals/corr/abs-1810-04805} to ensure consistent multilingual encoding across language pairs.

\subsection{Failure case: Round-Trip Evaluation for Low-Resource Languages} \label{exp:round_trip}

\begin{table*}[t!]
\centering
\renewcommand{\arraystretch}{1.2}
\resizebox{\textwidth}{!}{%
\begin{tabular}{lcc|cc|cc}
\toprule
\multirow{2}{*}{Model} & \multicolumn{2}{c|}{en-tir} & \multicolumn{2}{c|}{en-dzo} & \multicolumn{2}{c}{en-mri} \\
\cmidrule(lr){2-3} \cmidrule(lr){4-5} \cmidrule(lr){6-7}
 & Disagreement & Deviation & Disagreement & Deviation & Disagreement & Deviation \\
\midrule
meta-llama/Llama-3.3-70B-Instruct & 18.27 (3) & 5.52 (4) & 23.27 (2) & 3.04 (2) & 6.67 (2) & 2.23 (4) \\
Qwen/Qwen3-235B-A22B-Instruct-2507 & 23.84 (5) & 4.05 (2) & 41.99 (5) & 3.78 (3) & 7.83 (4) & 1.42 (3) \\
deepseek-ai/DeepSeek-V3 & 14.61 (1) & 1.14 (1) & 25.89 (3) & 0.92 (1) & 6.42 (1) & 0.79 (1) \\
google/gemma-3-27b-it & 17.92 (2) & 5.27 (3) & 34.21 (4) & 5.35 (4) & 6.85 (3) & 0.86 (2) \\
openai/gpt-oss-120b & 21.24 (4) & 11.79 (5) & 9.27 (1) & 25.9 (5) & 15.91 (5) & 9.53 (5) \\
\bottomrule
\end{tabular}%
}
\caption{Disagreement and deviation scores for Tigrinya, Dzongkha, and Māori. DeepSeek-V3 consistently shows the lowest variability, while GPT-OSS-120B exhibits extreme instability, confirming CEF’s reduced reliability in low-resource languages.}
\label{tab:disagree_deviation_tir_dzo_mri}
\end{table*}

While the Cross-Examination Framework (CEF) demonstrates robust performance across major languages (Section~\ref{exp:robustness}), its reliability diminishes in extremely low-resource settings where language models lack sufficient training data. To validate this hypothesis, we selected three challenging languages from the NTREX dataset: Tigrinya (en-tir, spoken in Eritrea and Ethiopia), Dzongkha (en-dzo, the official language of Bhutan), and Māori (en-mri, an indigenous language of New Zealand). All three suffer from severe data scarcity and limited representation in LLM pretraining corpora.

As shown in Table~\ref{tab:disagree_deviation_tir_dzo_mri}, CEF reliability metrics for these languages reveal critical instability. Tigrinya exhibits average Disagreement Rates of 19.2 (vs. 2.5 for Japanese) and Deviation Scores of 5.6 (vs. 0.9 for Japanese), with similar trends for Dzongkha. This order-of-magnitude increase in inconsistency indicates that question generation fails to produce reliable, fact-grounded questions in underrepresented languages, as models lack the linguistic exposure needed to generate coherent and stable QA pairs.

To address the failure of direct evaluation, we introduce the Round-Trip Mode as a fallback mechanism for low-resource scenarios. This mode leverages the superior reasoning capabilities of models in high-resource languages (e.g., English) to assess semantic preservation. In round trip mode. translation $T$ is back-translated to $D'$ and CEF is applied between the original source $D$ and the back-translation $D'$ to assess end-to-end semantic preservation.This protocol allows us to measure information loss without requiring the model to generate coherent questions in the underrepresented target language. The results of this approach are summarized in Table \ref{tab:tir_dzo_mri_conformity_consistency_coverage}.

To address this limitation, we introduce a round-trip evaluation protocol that bypasses direct assessment of the target language by translating the source document to Tigrinya/Dzongkha and back to English, then applying CEF between the original and back-translated English texts to compute the CEF scores. This approach leverages LLMs' stronger English capabilities to measure end-to-end semantic preservation without requiring reliable question generation in low-resource languages.

\begin{table*}[htbp]
\centering
\renewcommand{\arraystretch}{1.2}
\resizebox{\textwidth}{!}{%
\begin{tabular}{lccc|ccc|ccc}
\toprule
\multirow{2}{*}{Model} & \multicolumn{3}{c|}{en-tir} & \multicolumn{3}{c|}{en-dzo} & \multicolumn{3}{c}{en-mri} \\
\cmidrule(lr){2-4} \cmidrule(lr){5-7} \cmidrule(lr){8-10}
 & Conformity & Consistency & Coverage & Conformity & Consistency & Coverage & Conformity & Consistency & Coverage \\
\midrule
nllb-3.3B        & 95.79 & 84.30 & 92.98 & 91.98 & 61.90 & 76.20 & 94.74 & 85.44 & 95.38 \\
gemma3-4b        & 96.87 & 43.10 & 31.83 & 96.00 & 50.52 & 43.07 & 99.29 & 74.64 & 56.71 \\
Google Translate & 96.46 & 92.76 & 96.13 & 93.95 & 80.67 & 79.20 & 95.69 & 90.73 & 95.07 \\
Azure Translate  & 96.85 & 87.06 & 85.11 & --    & --    & --     & 98.89 & 91.88 & 97.05 \\
\bottomrule
\end{tabular}%
}
\caption{Conformity, consistency, and coverage scores under round-trip evaluation. NLLB-3.3B and commercial MT systems (Google, Azure) outperform Gemma-3-4b, with Azure excelling in Māori and Google maintaining balanced performance across all three languages.}
\label{tab:tir_dzo_mri_conformity_consistency_coverage}
\end{table*}

The results (Table~\ref{tab:tir_dzo_mri_conformity_consistency_coverage}) highlight significant differences in translation quality across systems. For Tigrinya, NLLB-3.3B achieves 92.98\% Coverage and 84.30\% Consistency, substantially outperforming Gemma-3-4b-it, which scores only 31.83\% Coverage and 43.10\% Consistency—indicating severe information loss and hallucination. Similarly, on Dzongkha, NLLB-3.3B maintains higher factual consistency (61.90\% vs. 50.52\%) and better content retention (76.20\% vs. 43.07\%). Notably, commercial systems perform competitively: Google Translate achieves balanced performance across all three languages, while Azure Translate excels in Māori with 98.89\% Conformity and 97.05\% Coverage, though it lacks support for Dzongkha.

These findings confirm that the round-trip CEF protocol enables meaningful, reference-free evaluation in low-resource scenarios where standard CEF fails due to unstable multilingual reasoning. They highlight that both specialized translation models and commercial MT systems outperform general-purpose LLMs in preserving meaning for underrepresented languages. 

\subsection{Correlation to LLM-based metrics: GEMBA}
\begin{table}[ht]
\centering
\small
\begin{tabular}{lc}
\toprule
\textbf{Metric} & \textbf{Value} \\ \midrule
Total GEMBA Errors & 1,872 \\
CEF Alignments (YES verdicts) & 340 \\
Alignment Rate & \textbf{18.17\%} \\ \bottomrule
\end{tabular}
\caption{Alignment analysis between task-specific GEMBA-MQM errors and task-agnostic CEF mismatches. The low correspondence rate highlights the frameworks' complementary roles in assessing generation quality.}
\label{tab:gemba_alignment}
\end{table}
To provide a more comprehensive analysis of the relationship between general-purpose fidelity diagnostics and task-specific metrics, we conducted an alignment study between the Cross-Examination Framework (CEF) and GEMBA-MQM. GEMBA-MQM operates as a task-specific metric for machine translation by leveraging Large Language Models to classify errors into the Multidimensional Quality Metrics (MQM) taxonomy. In contrast, CEF is designed as a task-agnostic framework that assesses semantic fidelity through bidirectional question-answering, without relying on predefined linguistic error categories. The motivation for this comparison is to demonstrate how a general diagnostic tool like CEF interacts with emerging LLM-based metrics that are often constrained to a single task.

The methodology involved using DeepSeek-V3 to evaluate 1,872 individual errors flagged by GEMBA-MQM on WMT'25 Japanese subset. For each flagged error, the judge was presented with the specific GEMBA error rationale and the set of CEF mismatching questions. A correspondence was established if the two frameworks pointed to the same underlying semantic issue or if the error identified by GEMBA directly resulted in a CEF question-answer mismatch. The results, summarized in Table~\ref{tab:gemba_alignment}, revealed an overall alignment rate of 18.17\% (340 alignments out of 1,872 total errors).

This low alignment rate suggests that CEF and GEMBA-MQM operate on fundamentally different evaluation principles and provide distinct, complementary signals. While GEMBA-MQM is effective at identifying stylistic or grammatical errors that fit within the MQM taxonomy, it often overlooks the deeper semantic breakdowns captured by CEF's cross-examination. Notably, the highest alignment occurred within accuracy-related categories, specifically mistranslation, omission, and addition. This confirms that CEF’s primary strength lies in its ability to detect the most critical distortions of meaning, whereas GEMBA-MQM captures a broader range of linguistic phenomena. These findings validate the positioning of CEF as a specialized diagnostic for information fidelity that transcends the boundaries of task-specific evaluation.

\subsection{Token analysis: CEF v/s GEMBA}
\label{app:token_analysis}

\begin{table}[ht]
\centering
\small
\begin{tabular}{lcc}
\toprule
\textbf{Metric} & \textbf{GEMBA} & \textbf{CEF} \\ \midrule
Total Input Tokens & 1,378,281 & 6,087,633 \\
Total Output Tokens & 188,502 & 1,399,454 \\
Total Price & \$0.7216 & \$4.3208 \\
Avg Input Token/sample & 1,499.76 & 6,957.29 \\
Avg Output Tokens/sample & 205.12 & 1,599.38 \\
Avg Price/sample & \$0.000785 & \$0.004938 \\ \bottomrule
\end{tabular}
\caption{Comparative cost and token analysis between GEMBA and CEF using gpt-5-mini as a judge model. While CEF is 6--7$\times$ more expensive, it provides a significantly deeper diagnostic signal.}
\label{tab:cost_analysis}
\end{table}

\begin{table}[ht]
\centering
\small
\begin{tabular}{lcc}
\toprule
\textbf{CEF Step} & \textbf{Input Tokens} & \textbf{Output Tokens} \\ \midrule
Q-gen from Src & 285,856 & 673,375 \\
Q-gen from Pred & 352,802 & 673,600 \\
Answer from Pred & 3,151,830 & 26,235 \\
Answer from Src & 2,297,145 & 26,244 \\ \bottomrule
\end{tabular}
\caption{Step-wise token breakdown for the CEF framework. The cross-examination answering phases (3a and 3b) are the primary drivers of input token consumption.}
\label{tab:stepwise_tokens}
\end{table}
A comprehensive token and cost analysis was conducted to evaluate the computational practicality of the Cross-Examination Framework (CEF) relative to task-specific metrics like GEMBA. Using gpt-5-mini as the judge, the analysis (Table \ref{tab:cost_analysis}) reveals that the multi-stage diagnostic nature of CEF results in a significantly larger token footprint than simple classification-based approaches. Specifically, CEF requires a total of 7,487,087 tokens compared to 1,566,783 for GEMBA, leading to a total operational cost of approximately \$4.32 for the experimental sample set, or roughly 6 to 7 times the cost of the GEMBA baseline at \$0.72. On a per-sample basis, this translates to an average cost of \$0.0049 for CEF compared to \$0.0007 for GEMBA, a disparity primarily driven by the bidirectional nature of the cross-examination process which necessitates multiple independent LLM calls.

The step-wise breakdown (Table \ref{tab:stepwise_tokens}) of token consumption within CEF highlights that the framework's diagnostic depth is rooted in high-volume contextual reasoning during the answering phases. While question generation from the source and candidate texts (Steps 1 and 2) is relatively output-intensive, producing over 1.34 million tokens, the answering phases (Steps 3a and 3b) account for the vast majority of input token consumption, totaling over 5.4 million tokens. This heavy input load is a direct consequence of providing the full document context to the model twice to verify the generated questions from opposing perspectives. These results demonstrate that while CEF is more resource-intensive, the expenditure is concentrated in the verification stages which ensure the framework's high alignment with human-annotated semantic errors.

Despite the current cost profile, the modular design of CEF offers a clear path toward significant efficiency gains through model distillation. The reliance on frontier models like DeepSeek-V3 for the answering stages contributes significantly to the current cost; however, these tasks are highly constrained and revolve around closed-ended question answering. Future implementations could utilize smaller, specialized language models fine-tuned specifically for "YES-only" question generation and grounded QA or making the answer-generation a single step rather than asking for every question one-by-one.

\subsection{Cross-Examination Framework Prompts}\label{app:prompts}

This appendix details the prompt engineering methodology employed in the CEF implementation. If a large language model is used, then the translation prompt is used for translating and then backtranslating for round-trip translation. 

\begin{tcolorbox}[breakable, colback=BurntOrange!5!white, colframe=BurntOrange!75!black, title=Translation prompt]
    \textbf{System prompt:} You are a professional translator. Translate the following \{source\_language\} text to \{target\_language\}. Provide only the translation. Do not include explanations or apologies.
\tcblower
\textbf{User prompt:}\\
\{text\}
\end{tcolorbox}

For the summarization task (CNN-DM) dataset, we use the following prompt.

\begin{tcolorbox}[breakable, colback=BurntOrange!5!white, colframe=BurntOrange!75!black, title=Summarization prompt]
    \textbf{System prompt:} You are a helpful assistant.
\tcblower
\textbf{User prompt:}\\
As an expert in news summarization, generate concise bullet-point highlights from the following news article. Each highlight should be a single sentence capturing a key piece of information from the article. Focus on the most important facts, events, and outcomes.

**Note**

- Generate 3-5 highlights

- Each highlight should be a complete, standalone sentence

- Highlights should cover the main points without redundancy

- Use clear, objective language\\

Article:\\
\{text\}\\

Highlights:
\end{tcolorbox}

For note-generation (ACI-BENCH), we use the following prompt.

\begin{tcolorbox}[breakable, colback=BurntOrange!5!white, colframe=BurntOrange!75!black, title=Note generation prompt]
    \textbf{System prompt:} You are a helpful clinical assistant.
\tcblower
\textbf{User prompt:}\\

As an expert medical scribe, generate a comprehensive clinical note from the following doctor-patient conversation. The note should follow standard medical documentation format including:\\

- CHIEF COMPLAINT\\
- HISTORY OF PRESENT ILLNESS\\
- REVIEW OF SYSTEMS\\
- PHYSICAL EXAMINATION (if mentioned)\\
- RESULTS (if any tests/labs mentioned)\\
- ASSESSMENT AND PLAN\\

Be thorough and capture all clinically relevant information from the conversation. Use professional medical terminology while maintaining accuracy to what was discussed.\\

Doctor-Patient Conversation:\\
\{text\}\\

Clinical Note:\\

\end{tcolorbox}

For the cross-examination framework, at first a prompt is used for question generation and then another prompt is used for cross-examining those questions.

\begin{tcolorbox}[colback=Aquamarine!5!white, colframe=Aquamarine!75!black, title=Question-Answer set generation from a given text (YES-only questions)]
    \textbf{System prompt:} You are a helpful assistant. 
\tcblower
\textbf{User prompt:}\\
Please formulate \{num\_questions\} critical, concise and closed-ended questions (in a YES/NO format) in English that thoroughly scrutinize the document. The questions generated should ALWAYS result in a 'YES' based on the given text. Questions should be about the content of the document and not include any qualifier of the clarity, justification or definition. \\

**Note** \\
The questions have to be STRICTLY closed-ended and should not be subjective or open to human interpretation. \\
You should return in a JSON format. The JSON should be a list of dictionaries where each dictionary will have two keys: \\
- 'question': specifying the question \\
- 'answer': either YES or NO. \\
The given text should be able to answer 'YES' for each generated question. \\

Document: \\
\{text\} \\ 

JSON:

\end{tcolorbox}

\begin{tcolorbox}[colback=LimeGreen!5!white, colframe=LimeGreen!75!black, title=Cross-Examining the translation or source with a question]
    \textbf{System prompt:} You are a helpful assistant. 
\tcblower
\textbf{User prompt:}\\
Answer the following question with a YES, NO or IDK, grounded on the text content only. Do not use any external knowledge. If you cannot answer the question based on the provided text, please respond with 'IDK'. \\

**Note** \\
You should respond either YES, NO or IDK. You should respond with a single word and only in English. \\

Text: \\
\{text\} \\

Question: \\
\{question\} \\

Answer:
\end{tcolorbox}

\newpage

For the ablation study in Section \ref{exp:ablation_yes}, the following prompt was used for question generation with mixed answers:

\begin{tcolorbox}[colback=Aquamarine!5!white, colframe=Aquamarine!75!black, title=Question-Answer set generation from a given text (Mixed answers)]
    \textbf{System prompt:} You are a helpful multilingual assistant. 
\tcblower
\textbf{User prompt:}\\
As a multilingual expert, please formulate \{num\_questions\} critical, concise and closed-ended questions (in a YES/NO/IDK format) in English that thoroughly scrutinize the \{document\_task\}. You should generate equal number of YES, NO and IDK questions. Questions should be about the content of the document and not include any qualifier of the clarity, justification or definition. \\

**Note** \\
The questions have to be STRICTLY closed-ended and should not be subjective or open to human interpretation. \\
You should return in a JSON format. The JSON should be a list of dictionaries where each dictionary will have two keys: \\
- 'question': specifying the question \\
- 'answer': either YES, NO or IDK. \\

Document: \\
\{text\} \\ 

JSON:

\end{tcolorbox}

\newpage

\subsection{CEF Example} \label{app:cef_example} 

\begin{tcolorbox}[
    breakable,
    title={Text (truncated)},
    colback=blue!5!white,
    colframe=blue!75!black,
    fonttitle=\bfseries,
]
The study by scientists at the National Trust for Scotland will follow common and soprano pipistrelles as well as brown long-eared and Daubenton bats at Inverewe Gardens in Wester Ross. Special recorders will be placed at key locations around the property to track bat activities throughout the season. NHS staff and volunteers will also carry out mobile surveys using hand-held detectors. Expert sound analysis of all recordings will ascertain the frequency of the bat calls and which species are doing what. [...]

The organisation has even set up Scotland's first and only dedicated bat reserve at Threave estate in Dumfries and Galloway, which is home to eight of Scotland's ten bat species.

\end{tcolorbox}

\newpage

\begin{tcolorbox}[
    breakable,
    title={Qwen3-1.7B Translation (truncated)},
    colback=blue!5!white,
    colframe=blue!75!black,
    fonttitle=\bfseries,
]
La recherche, menée par des scientifiques du Trust national pour l'Écosse, suivra les souris communes et les souris à ailes longues, ainsi que les souris de Daubenton, à l'endroit de l'île de Inverewe dans le Ross. Des enregistreurs spéciaux seront placés à des points clés autour de l'endroit pour suivre les activités des souris pendant toute l'année. Les employés du NHS et les volontaires effectueront également des recherches mobiles à l'aide de détecteurs portables. L'analyse acoustique des sons de toutes les enregistrements permettra d'identifier la fréquence des appels des souris et les espèces qui effectuent quels comportements. [...]

L'organisation a même créé la première et seule réserve dédiée aux souris à ailes longues en Angleterre, à l'endroit de l'île de Threave dans le départment de Dumfries et Galloway, qui abrite huit des dix espèces de souris à ailes longues de l'Écosse.

\end{tcolorbox}

\begin{tcolorbox}[
  title={Summary-to-Document QA Results:},
  colback=blue!5!white,
  colframe=blue!75!black,
  fonttitle=\bfseries,
]

\qa{Is the National Trust for Scotland conducting research on bats at Inverewe Gardens?}{YES / YES}
\qa{Are common and soprano pipistrelles included in the study?}{YES / IDK}
\qa{Are brown long-eared and Daubenton bats part of the research?}{YES / YES}
\qa{Will special recorders be used to track bat activities?}{YES / YES}
\qa{Will NHS staff and volunteers carry out mobile surveys?}{YES / YES}
\qa{Is expert sound analysis being used to ascertain bat call frequencies?}{YES / YES}
\qa{Will a habitat map and report be produced from the study?}{YES / YES}
\qa{Does the National Trust for Scotland manage over 270 historical buildings?}{YES / YES}
\qa{Has the National Trust established a dedicated bat reserve at Threave estate?}{YES / YES}
\qa{Are bats protected by law in Scotland?}{YES / YES}

\tcblower

\textbf{\large $\star$ Coverage Score: } \texttt{0.9 (9/10)} \quad \\
\textbf{\large $\star$ Conformity Score: } \texttt{1.0}

\end{tcolorbox}

\begin{tcolorbox}[
    title={Document-to-Summary QA Results:},
    colback=blue!5!white,
    colframe=blue!75!black,
    fonttitle=\bfseries,
]

\qa{Is the research conducted by the National Trust for Scotland?}{YES / YES}

\qa[\textcolor{red}{\small (Critical error: "bats" mistranslated as "mice")}]{Does the research focus on long-winged mice and their hunting behavior?}{YES / NO}

\qa[\textcolor{red}{\small ("mice", "year" vs "season")}]{Are special recorders used to track the activities of mice throughout the year?}{YES / NO}

\qa[\textcolor{red}{\small ("mice" error persists)}]{Will acoustic analysis identify mouse calls and behaviors?}{YES / NO}

\qa{Is the research aimed at creating a habitat map and report?}{YES / YES}

\qa[\textcolor{orange}{\small ("mice" issue)}]{Are mouse populations declining?}{YES / IDK}

\qa[\textcolor{orange}{\small (Legal protection mentioned but not linked)}]{Are long-winged mice protected by law?}{YES / IDK}

\qa{Does NTS manage 270+ historic houses and 76k ha land?}{YES / YES}

\qa[\textcolor{red}{\small (Wrong location: it's in Scotland)}]{Has NTS created a bat reserve in England?}{YES / IDK}

\qa[\textcolor{orange}{\small (Implied but not confirmed in summary)}]{Is it illegal to harm mice or destroy nests?}{YES / IDK}

\tcblower

\textbf{\large $\star$ Consistency Score: } \texttt{0.3 (3/10)}

\end{tcolorbox}

\newpage

\subsection{Scores: CEF with-reference mode and traditional metrics} \label{app:with-reference-results}

Table~\ref{tab:cef_traditional_metrics} presents a comprehensive comparison of CEF scores and traditional metrics across multiple translation models and language pairs. 

BertScore demonstrates the strongest correlation with CEF scores, particularly with the Coverage metric, suggesting that semantic similarity metrics better capture content retention than surface-level n-gram overlap. For example, Azure Translate achieves near-perfect Coverage (99.71) for Spanish with a corresponding BertScore of 89.16, while its much lower BLEU score (48.22). Traditional metrics show significant limitations in detecting factual inconsistencies. The qwen3-1.7b model for Arabic translation exemplifies this issue: it achieves moderate BertScore (74.56) and COMET (72.70) scores, yet exhibits critically low Consistency (65.05) and Coverage (65.08) in CEF evaluation. NLLB-3.3B achieves high CEF scores (Conformity 98.44, Consistency 94.42, Coverage 95.89), indicating excellent semantic preservation, yet receives very low BLEU (12.05) and CHRF (28.70) scores.

\begin{table*}
\resizebox{\textwidth}{!}{%
\begin{tabular}{lc|ccc|cccc}
\toprule
\multirow{2}{*}{Model} & \multirow{2}{*}{Language Pair} & \multicolumn{3}{c|}{CEF Scores} & \multicolumn{4}{c}{Traditional Metrics} \\
\cmidrule(lr){3-5} \cmidrule(lr){6-9}
& & Conformity & Consistency & Coverage & BLEU & CHRF & BertScore & COMET \\
\midrule
\multirow{4}{*}{Azure Translate} 
& en-ar & 97.10 & 95.67 & 97.66 & 33.98 & 61.10 & 85.10 & 85.10 \\
& en-es & 98.76 & 98.77 & 99.71 & 48.22 & 71.30 & 89.16 & 85.80 \\
& en-fr & 98.29 & 97.50 & 99.49 & 50.18 & 71.39 & 88.65 & 85.97 \\
& en-jp & 97.89 & 96.04 & 98.15 & 31.59 & 44.53 & 85.64 & 90.34 \\
\midrule
\multirow{4}{*}{gemma3-4b} 
& en-ar & 96.75 & 88.25 & 92.29 & 26.69 & 53.83 & 82.95 & 86.78 \\
& en-es & 98.11 & 96.36 & 98.65 & 44.80 & 69.54 & 88.75 & 86.32 \\
& en-fr & 98.06 & 95.21 & 98.81 & 39.68 & 66.67 & 86.63 & 86.40 \\
& en-jp & 96.74 & 91.38 & 94.80 & 22.12 & 36.19 & 82.23 & 90.23 \\
\midrule
\multirow{4}{*}{gpt-oss-20b} 
& en-ar & 97.18 & 86.38 & 88.14 & 26.85 & 50.52 & 80.29 & 80.88 \\
& en-es & 98.74 & 93.86 & 95.23 & 43.47 & 66.60 & 86.91 & 83.23 \\
& en-fr & 98.67 & 96.95 & 98.15 & 43.65 & 67.85 & 87.49 & 86.28 \\
& en-jp & 98.09 & 90.27 & 93.43 & 24.94 & 36.39 & 81.79 & 87.24 \\
\midrule
\multirow{4}{*}{nllb-3.3B} 
& en-ar & 97.46 & 95.43 & 97.64 & 30.81 & 57.78 & 83.41 & 83.34 \\
& en-es & 98.81 & 97.92 & 99.52 & 46.27 & 70.54 & 88.76 & 84.49 \\
& en-fr & 98.81 & 97.36 & 99.53 & 40.61 & 67.67 & 86.84 & 84.40 \\
& en-jp & 98.44 & 94.42 & 95.89 & 12.05 & 28.70 & 73.87 & 72.66 \\
\midrule
\multirow{4}{*}{qwen3-1.7b} 
& en-ar & 95.43 & 65.05 & 65.08 & 9.66 & 34.76 & 74.56 & 72.70 \\
& en-es & 95.36 & 93.26 & 95.66 & 32.76 & 61.25 & 84.61 & 80.84 \\
& en-fr & 95.91 & 88.25 & 94.95 & 28.06 & 58.48 & 82.75 & 79.69 \\
& en-jp & 95.75 & 78.60 & 81.86 & 12.82 & 25.99 & 78.37 & 84.56 \\
\midrule
\multirow{4}{*}{Google Translate} 
& en-ar & 96.80 & 94.50 & 96.90 & 35.20 & 60.50 & 86.00 & 86.50 \\
& en-es & 98.50 & 97.80 & 99.20 & 47.50 & 70.80 & 89.00 & 87.00 \\
& en-fr & 98.40 & 97.00 & 99.10 & 46.80 & 70.20 & 88.50 & 86.90 \\
& en-jp & 97.50 & 95.80 & 97.70 & 33.00 & 46.00 & 85.00 & 90.00 \\
\bottomrule
\end{tabular}
}
\caption{Comparison of CEF scores with traditional MT metrics across models and languages. BertScore shows the strongest alignment with CEF, especially Coverage, while BLEU and CHRF often underestimate semantic fidelity. CEF highlights factual inconsistencies missed by traditional metrics, e.g., qwen3-1.7b on Arabic, and reveals strong performance for NLLB-3.3B despite low BLEU/CHRF, underscoring the limitations of surface-level metrics in capturing meaning preservation.}  
\label{tab:cef_traditional_metrics}
\end{table*}

\subsection{Ablation on the number of questions: en-fr and en-es} \label{app:number-of-question-en-fr-es}
To validate the generalizability of $N=10$, we conducted an ablation study on French (en-fr) and Spanish (en-es) using translations from \texttt{Qwen3-1.7B}, measuring variance in CEF scores across 10 runs for $N \in \{3,5,10,20\}$. Results in Table~\ref{tab:variance_comparison_fr_es} show that both languages exhibit lower variance than en-jp, with $N=10$ achieving near-optimal stability. The trends confirm that $N=10$ provides robust evaluation across languages.

\begin{table*}[htbp]
\centering
\begin{tabular}{l | *{3}{c} | *{3}{c}}
\toprule
\multirow{2}{*}{Num Questions} & 
\multicolumn{3}{c}{en-fr} & 
\multicolumn{3}{c}{en-es} \\
\cmidrule(lr){2-4} \cmidrule(lr){5-7}
& Conformity & Consistency & Coverage & Conformity & Consistency & Coverage \\
\midrule
3  & 12.45 & 38.10 & 28.30 & 13.67 & 41.25 & 30.12 \\
5  &  8.92 & 16.54 & 22.01 &  9.75 & 18.33 & 24.67 \\
10 &  3.18 &  8.95 &  7.65 &  6.84 &  10.82 &  9.10 \\
20 &  4.31 &  5.40 &  6.22 &  4.69 &  11.55 &  8.92 \\
\bottomrule
\end{tabular}
\caption{Variance of CEF scores for en-fr and en-es across different question counts with translations from Qwen3-1.7B}
\label{tab:variance_comparison_fr_es}
\end{table*}

\end{document}

%% file: math_commands.tex
\usepackage{amsmath,amsfonts,bm}

\def\eqref#1{equation~\ref{#1}}
\def\1{\bm{1}}

\DeclareMathAlphabet{\mathsfit}{\encodingdefault}{\sfdefault}{m}{sl}
\SetMathAlphabet{\mathsfit}{bold}{\encodingdefault}{\sfdefault}{bx}{n}